# A Comparative Study of Fuzzy Classification Methods on Breast Cancer Data[*]

Ravi. Jain[1], Ajith. Abraham[2]

[1]School of Computer & Information Science,
University of South Australia,
Mawson Lakes Boulevard, Mawson Lakes, SA 5095
ravi.jain@unisa.edu.au

[2]Computer Science Department,
Oklahoma State University, 700 N Greenwood Avenue, Tulsa, Oklahoma OK 74106, USA.
ajith.abraham@ieee.org

## Abstract

In this paper, we examine the performance of four fuzzy rule generation methods on Wisconsin breast cancer data. The first method generates fuzzy if-then rules using the mean and the standard deviation of attribute values. The second approach generates fuzzy if-then rules using the histogram of attributes values. The third procedure generates fuzzy if-then rules with certainty of each attribute into homogeneous fuzzy sets. In the fourth approach, only overlapping areas are partitioned. The first two approaches generate a single fuzzy if-then rule for each class by specifying the membership function of each antecedent fuzzy set using the information about attribute values of training patterns. The other two approaches are based on fuzzy grids with homogeneous fuzzy partitions of each attribute. The performance of each approach is evaluated on breast cancer data sets. Simulation results show that the Modified grid approach has a high classification rate of 99.73 %.

## 1. Introduction

Breast cancer is the most common cancer in women in many countries. Most breast cancers are detected as a lump/mass on the breast, or through self-examination/mammography [1]. Screening mammography is the best tool available for detecting cancerous lesions before clinical symptoms appear. Surgery through a biopsy or lumpectomy have been also been the most common methods of removal. Fine needle aspiration (FNA) of breast masses is a cost-effective, non-traumatic, and mostly invasive diagnostic test that obtains information needed to evaluate malignancy. Recently, a new less invasive technique, which uses super-cooled nitrogen to freeze and shrink a non-cancerous tumor and destroy the blood vessels feeding the growth of the tumour, has been developed [2] in the USA. Several Artificial Intelligence (AI) techniques including neural networks and fuzzy logic [3-5] are successfully applied to a wide variety of decision-making problems in the area of medical diagnosis. In this paper we examine the performance of four direct rule generation methods that involve no time-consuming tuning procedures on *breast cancer data* [6]. The first method generates fuzzy if-then rules using the mean and the standard deviation of attribute values. The second approach generates fuzzy if-then rules using the histogram of attributes values. The third approach generates fuzzy if-then rules with certainty of each attribute into homogeneous fuzzy sets. In the fourth approach, only overlapping areas are partitioned. This approach is a modified version of the third approach. In the first two approaches, a single fuzzy if-then rule is generated for each class. That is, the number of fuzzy if-then rules is the same as the number of classes. These methods were reported in [7-8]. The main advantage of fuzzy rule-based systems is that they do not require large memory storage, their inference speed is very high and the users can carefully examine each fuzzy if-then rule. This paper is organised as follows. Section 2 describes the characteristics of fuzzy systems. Section 3 describes rule generation methods. Section 4 provides details of the Wisconsin breast cancer data, and Section 5 describes simulation and results. The final section provides some conclusions relating to the performance of fuzzy systems when applied to the breast cancer data.

## 2. Fuzzy Systems

Fuzzy logic was invented by Zadeh [9] in 1965 for handling uncertain and imprecise knowledge in real world applications. It has proved to be a powerful tool for decision-making, and to handle and manipulate imprecise and noisy data. The first major commercial application was in the area of cement kiln control. This requires that an

---

operator monitor four internal states of the kiln, control four sets of operations, and dynamically manage 40 or 50 "rules of thumb" about their interrelationships, all with the goal of controlling a highly complex set of chemical interactions. One such rule is "If the oxygen percentage is rather high and the free-lime and kiln-drive torque rate is normal, decrease the flow of gas and slightly reduce the fuel rate".

The notion central to fuzzy systems is that truth values (in fuzzy logic) or membership values (in fuzzy sets) are indicated by a value on the range [0.0, 1.0], with 0.0 representing absolute Falseness and 1.0 representing absolute Truth. A fuzzy system is characterized by a set of linguistic statements based on expert knowledge. The expert knowledge is usually in the form of "*if-then*" rules.

*Definition 1*: A fuzzy set A in X is characterized by a membership function which is easily implemented by fuzzy conditional statements. For example, if the antecedent is true to some degree of membership, then the consequent is also true to that same degree.

*If* antecedent **Then** consequent

*Rule:* **If** *variable₁* is *low* and *variable₂* is *high* **Then** *output* is *benign* **Else** *output* is *malignant*

In a fuzzy classification system, a case or an object can be classified by applying a set of fuzzy rules based on the linguistic values of its attributes.

Every rule has a weight, which is a number between 0 and 1, and this is applied to the number given by the antecedent. It involves 2 distinct parts. The first part involves evaluating the antecedent, fuzzifying the input and applying any necessary fuzzy operators. For Example,

Union: $\mu_{A \cap B}(x) = \text{Min}[\mu_A(x), \mu_B(x)]$

Intersection: $\mu_{A \cap B}(x) = \text{Min}[\mu_A(x), \mu_B(x)]$

Complement: $\mu_{\overline{A}}(x) = 1 - \mu_A(x)$

where $\mu$ is the membership function.

The second part requires application of that result to the consequent, known as inference. To build a fuzzy classification system, the most difficult task is to find a set of fuzzy rules pertaining to the specific classification problem.

A fuzzy inference system is a rule-based system that uses fuzzy logic, rather than Boolean logic, to reason about data. Its basic structure includes four main components (1) a fuzzifier, which translates crisp (real-valued) inputs into fuzzy values; (2) an inference engine that applies a fuzzy reasoning mechanism to obtain a fuzzy output; (3) a defuzzifier, which translates this latter output into a crisp value; and (4) a knowledge base, which contains both an ensemble of fuzzy rules, known as the rule base, and an ensemble of membership functions known as the database.

The decision-making process is performed by the inference engine using the rules contained in the rule base. These fuzzy rules define the connection between input and output fuzzy variables.

# 3. Rule Generation Procedure

In this section, we explain each of four approaches examined in this paper. The performance of each approach is examined in the next section by computer simulations on breast cancer data sets.

Let us assume that we have an n-dimensional c-class pattern classification problem whose pattern space is an n-dimensional unit cube $[0,1]^n$. We also assume that m patterns $x_p = (x_{p1},...,x_{pn})$, $p = 1,2,...,m$, are given for generating fuzzy if-then rules where $x_{pi} \in [0,1]$ for p =1,2,..., m, i =1,2,...,n. In computer simulations of this paper, all attribute values are normalized into the unit interval [0,1].

## 3.1 Rule Generation Based on the Mean and the Standard Deviation of Attribute Values

In this approach, a single fuzzy if-then rule is generated for each class. The fuzzy if-then rule for the $k^{th}$ class can be written as

If $x_1$ is $A_1$ and ... and $x_n$ is $A_n$ then Class k, (1)

where $A_i$ is an antecedent fuzzy set for the $i^{th}$ attribute. The membership function of $A_i$ is specified as

$$A_i(x_i) = \exp\left(-\frac{(x_i - \mu_i)^2}{2(\sigma_i)^2}\right) \quad (2)$$

where $\mu_i$ is the mean of the $i^{th}$ attribute values $x_{pi}$ of Class k patterns, and $\sigma_i$ is the standard deviation. Fuzzy if-then rules for the two-dimensional two-class pattern classification problem are written as follows:

The membership function of each antecedent fuzzy set is specified by the mean and the standard deviation of attribute values (see Figure 1). For a new pattern $x_p = (x_{p3}, x_{p4})$, the winner rule is determined as follows:

$$A_3^*(x_{p3}).A_2^*(x_{p4}) = \max\left\{A_1^k(x_{p3}).A_2^k(x_{p4}) \big| k = 1,2\right\} \quad (3)$$

For each attribute, 20 membership functions $f_h()$, h=1,2,...,20 were used. The fuzzy partition was used only for calculating the histogram.

## 3.2 Rule Generation Based on the Histogram of Attribute Values

In this method the use of histogram an antecedent membership function and each attribute is partitioned into several fuzzy sets. We used 20 membership functions $f_h(.)$, h=1,2,...,20 for each attribute in computer simulations as shown in Figure 2.

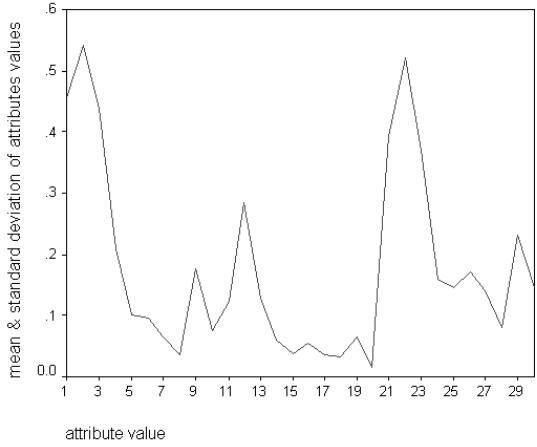

Figure 1: Mean and standard deviation of attributes values

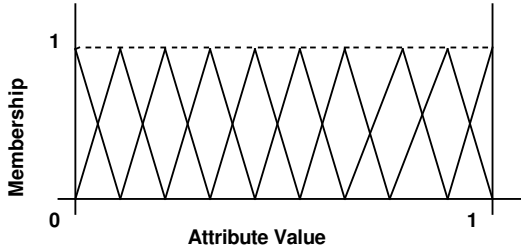

Figure 2: Fuzzy partition for calculating the smoothed histogram

The smoothed histogram $m_i^k(x_i)$ of Class k patterns for the $i^{th}$ attribute is calculated using the 20 membership functions $f_h(.)$ as follows:

$$m_i^k(x_i) = \frac{1}{m^k} \sum_{x_p \in Class\ k} f_h(x_{pi}) \quad (4)$$

$$\text{for } \beta_{h-1} \leq x_i \leq \beta_h, \ h=1,2,\ldots,20$$

where $m_k$ is the number of Class k patterns, $\left[\beta_{h-1}, \beta_h\right]$ is the $h^{th}$ crisp interval corresponding to the 0.5-level set of the membership function $f_h(.)$:

$$\beta_1 = 0, \ \beta_{20} = 1, \quad (5)$$

$$\beta_h = \frac{1}{20-1}\left(h - \frac{1}{2}\right) \text{ for } h=1,2,\ldots,19 \quad (6)$$

The smoothed histogram in (4) is normalized so that its maximum value is 1. An example of such a normalized histogram is shown in Figure 3, which is the histogram of Class 2 patterns for the $3^{rd}$ attribute of breast cancer data. As in the first approach based on the mean and the standard deviation, a single fuzzy *if-then* rule in (2) is generated for each class in the second approach.

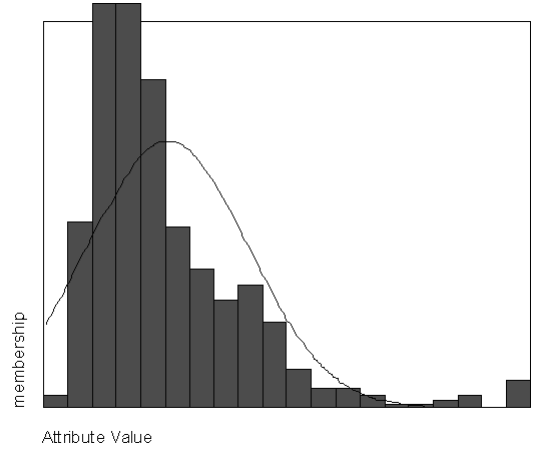

Figure 3: Normalised histogram of class2 patterns

## 3.3 Rule Generation of Based on Simple Fuzzy Grid

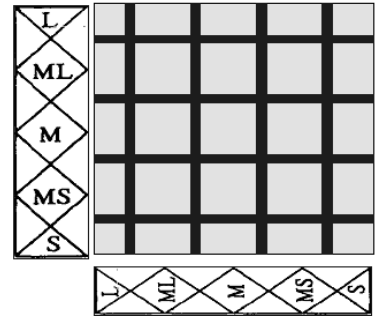

Figure 4: An example of fuzzy partition

In the first two approaches, a single fuzzy *if-then* rule was generated for each class using the information about training patterns. On the contrary, many fuzzy *if-then* rules are generated in the third approach by partitioning each attribute into homogeneous fuzzy sets. In Figure 4, we show an example of such a fuzzy partition where the two dimensional pattern space is partitioned into 25 fuzzy subspaces by five fuzzy sets for each attribute (S: small, MS: medium small, M: medium, ML: medium large, L: large). A single fuzzy *if-then* rule is generated for each fuzzy subspace. Thus, the number of possible fuzzy *if-then* rules in Figure 4 is 25.

One disadvantage of this approach is that the number of possible fuzzy *if-then* rules exponentially increases with the dimensionality of the pattern space. For coping with this difficulty, some GA-based rule selection approaches have been proposed to find a compact rule set [11]. The number of fuzzy *if-then* rules can be also decreased by feature selection [12].

Because the specification of each membership function does not depend on any information about training patterns, this approach uses fuzzy *if-then* rules with certainty grades. The local information about training patterns in the corresponding fuzzy subspace is used for determining the consequent class and the grade of certainty.

In this approach, fuzzy *if-then* rules of the following type are used:

**If** $x_1$ is $A_{j1}$ and ... and $x_n$ is $A_{jn}$

**Then**

Class $C_j$, with CF = $CF_j$, j = 1, 2, ..., N     (7)

where *j* indexes the number of rules, *N* is the total number of rules, $A_{ji}$ is the antecedent fuzzy set of the i$^{th}$ rule for the i$^{th}$ attribute, $C_j$; is the consequent class, and $CF_j$ is the grade of certainty. The consequent class and the grade of certainty of each rule are determined by the following simple heuristic procedure:

**Step 1:** Calculate the compatibility of each training pattern $x_p = (x_{p1}, x_{p2}, ..., x_{pn})$ with the j-th fuzzy if-then rule by the following product operation:

$$\pi_j(x_p) = A_{j1}(x_{p1}) \times ... \times A_{jn}(x_{pn}), \quad p = 1, 2, ..., m. \quad (8)$$

**Step 2:** For each class, calculate the sum of the compatibility grades of the training patterns with the j-th fuzzy if-then rule $R_j$:

$$\beta_{class\ k}(R_j) = \sum_{x_p \in class\ k}^{n} \pi(x_p), \quad k=1,2,...,c \quad (9)$$

where $\beta_{class\ k}(R_j)$ the sum of the compatibility grades of the training patterns in Class *k* with the j-th fuzzy if-then rule $R_j$.

**Step 3:** Find Class $A_j^*$ that has the maximum value $\beta_{class\ k}(R_j)$:

$$\beta_{class\ k_j^*} = Max\{\beta_{class\ 1}(R_j), ..., \beta_{class\ c}(R_j)\} \quad (10)$$

If two or more classes take the maximum value or no training pattern compatible with the j-th fuzzy if-then rule (i. e., if $\beta_{Class\ k}(R_j)=0$ for k =1,2,..., c ), the consequent class $C_i$ can not be determined uniquely. In this case, let $C_i$ be $\phi$. If a single class takes the maximum value, the consequent class $C_j$ *is* determined by (7).

**Step 4:** If the consequent class $C_i$ is 0, let the grade of certainty $CF_j$ be $CF_j = 0$. Otherwise the grade of certainty $CF_j$ *is* determined as follows:

$$CF_j = \frac{(\beta_{class\ k_j^*}(R_j) - \bar{\beta})}{\sum_{k=1}^{c} \beta_{class\ k}(R_j)} \quad (11)$$

where

$$\bar{\beta} = \sum_{\substack{k=1 \\ k \neq k_j^*}} \frac{\beta_{Class\ k}(R_j)}{(c-1)}$$

## 3.4. Rule Generation Based on Fuzzy Partition of Overlapping Areas

In the third approach, the shape of each membership function was specified without utilizing any information about training patterns. A simple modification of the third approach is to partition only overlapping areas. This approach is illustrated in Figure 5.

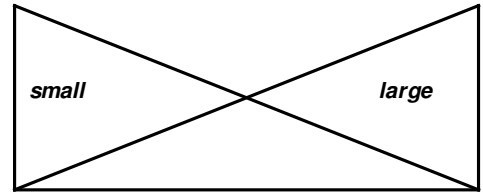

Attribute value

**(a)** Simple fuzzy grid approach

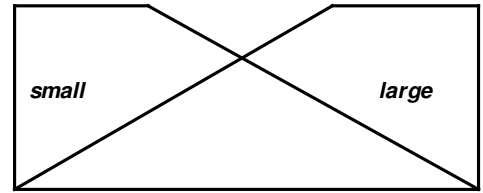

Overlapping area

**(b)** Modified fuzzy grid approach

**Figure 5: Fuzzy partition of each attribute**

This approach generates fuzzy *if-then* rules in the same manner as the simple fuzzy grid approach except for the specification of each membership function. Because this approach utilizes the information about training patterns for specifying each membership function as in the first and second approaches, the performance of generated fuzzy *if-then* rules is good even when we do not use the certainty grade of each rule in the classification phase. For example, the classification boundary in Figure 5 was obtained by generating nine fuzzy *if then* rules without certainty grades.

In this approach, the effect of introducing the certainty grade to each rule is not large when compared with the third approach. In computer simulations of the next section, we used fuzzy *if-then* rules with certainty grades in this approach, as in the third approach.

# 4. Wisconsin Diagnostic Breast Cancer Data

The Wisconsin breast cancer dataset [6] was obtained from a repository of a machine-learning database University of California, Irvine. This data set has 32 attributes (30 real-valued input features) and 569 instances of which 357 are of benign and 212 are of malignant class. Table 1 shows the statistical details of the data.

**Table 1: Statistical details of the data**

| Class | Frequency | Percent | Valid Percent | Cumulative Percent |
|---|---|---|---|---|
| 1 | 357 | 62.7 | 62.7 | 62.7 |
| 2 | 212 | 37.3 | 37.3 | 100.0 |
| Total | 569 | 100.0 | 100.0 | |

Several studies have been conducted based on this database. For example, Bennet and Mangasarian [10] used linear programming techniques, obtaining a 99.6% classification rate on 487 cases (the reduced database available at the time). However, diagnostic decisions are essentially black boxes, with no explanation as to how they were attained. Figure 6 shows 3D plot of data.

Ten real-valued features are computed for each cell nucleus:

a) radius (mean of distances from center to points on the perimeter)

b) texture (standard deviation of gray-scale values)

c) perimeter

d) area

e) smoothness (local variation in radius lengths)

f) compactness (perimeter^2 / area - 1.0)

g) concavity (severity of concave portions of the contour)

h) concave points (number of concave portions of the contour)

i) symmetry

j) fractal dimension ("coastline approximation" - 1)

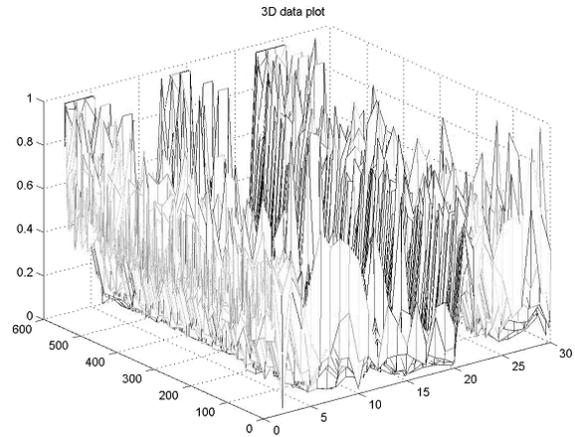

**Figure 6: 3D plot of Wisconsin data**

# 5. Simulation Results and Discussions

We examined the performance of four different approaches, and the empirical results are summarized in Table 2.

**Table 2. Classification rates for breast cancer data**

| | |
|---|---|
| Mean and Standard Deviation | **92.2%** |
| Histogram | 86.7% |
| Simple Grid | 99.73% |
| Modified Grid | 62.57% |

As evident, the performance of simple grid and mean and standard deviation is comparable. But the performance of histogram and modified grid approaches is not good enough with the other approaches. This is because in the histogram approach a single fuzzy rule is not enough for each class and the classification of some patterns was rejected and in the case of the modified grid approach the number of fuzzy if-then rules is increased exponentially with the dimensionality of pattern space. Simple grid approach gave the overall best results with a classification accuracy of 99.73%. Rule generation using mean and standard deviation is easy to implement as it depends only on the mean and standard deviation of the attribute values.

The modified grid approach did not produce the desired accuracy. Moreover, in the grid-based approach, the number of fuzzy if-then rules exponentially increased with the dimensionality of the pattern space. Thus, a large number of fuzzy if-then rules are usually generated for real-world pattern classification problems. This leads to several drawbacks: over-fitting training patterns, large memory storage requirement, and slow inference speed. On the contrary, the numbers of fuzzy if-then rules in the first two approaches are the same as the number of classes.

## 6. Conclusion and Discussions

In this paper, we examined the performance of four fuzzy rule generation methods that could generate fuzzy if-then rules directly from training patterns with no time-consuming tuning procedures. In the first approach, a single fuzzy if-then rule was generated for each class using the mean and the standard deviation of attribute values. In the second approach, a single fuzzy if-then rule was generated for each class using the histogram of attribute values. The third approach generated fuzzy if-then rules by homogeneously partitioning each attribute. Thus, a pattern space was partitioned into a simple fuzzy grid. The information about attribute values was not used for specifying the membership function of each antecedent fuzzy set. The local information of training patterns was utilized when the consequent class and the certainty grade were specified. The last approach was a modified version of the simple fuzzy grid approach.

As illustrated in Table 2, simple grid approach gave the best performance overall while the mean and standard deviation approach also performed reasonably well.

It may be noted that a single fuzzy if-then rule for each class is not always sufficient for real-world pattern classification problems. While each approach is very simple and has some drawbacks as discussed above, fuzzy rule-based systems have high classification ability as shown in this paper. The performance of fuzzy rule based systems can be further improved by feature selection and optimizing the rule selection and various rule parameters.

## Acknowledgements

The authors are grateful to the editor of this journal, John Pattison, reviewers (for constructive comments) and a number of readers including Robyn Vast for reading and correcting this paper.